\documentclass[10pt,journal,compsoc]{IEEEtran}
\usepackage{enumitem}
\usepackage{multirow}
\usepackage[font=normalsize]{subfig}
\usepackage{makecell}
\usepackage{amsmath,amsfonts,amsthm}
\usepackage{mathtools, nccmath}
\usepackage{array}
\usepackage{longtable}
\usepackage{rotating}
\usepackage{amssymb,xcolor,stackengine,graphicx}

\usepackage{url}            
\usepackage{booktabs}       

\ifCLASSOPTIONcompsoc
  \usepackage[nocompress]{cite}
\else
  \usepackage{cite}
\fi
\ifCLASSINFOpdf
\else
\fi

\begin{document}
%
\title{Mitigating Relational Bias on Knowledge Graphs}
%
%
%
%

\author{Yu-Neng Chuang, Kwei-Herng Lai, Ruixiang Tang, Mengnan Du, Chia-Yuan Chang, Na Zou and Xia Hu 
\IEEEcompsocitemizethanks{\IEEEcompsocthanksitem Yu-Neng Chuang, Kwei-Herng Lai, Ruixiang Tang and Xia Hu are with the Department of Computer Science, Rice University. 
E-mail: \{ynchuang, khlai, rt39, xia.hu\}@rice.edu.
\IEEEcompsocthanksitem Mengnan Du and Chia-Yuan Chang are with the Department of Computer Science and Engineering, Texas A\&M University.
Email: \{dumengnan, cychang\}@tamu.edu.
\IEEEcompsocthanksitem Na Zou is with the Department of Engineering Technology and Industrial Distribution, Texas A\&M University.
Email: nzou1@tamu.edu.
\IEEEcompsocthanksitem Correspondence to Yu-Neng Chuang and Xia Hu.}}

\markboth{UNDER REVIEW IEEE TRANSACTIONS ON PATTERN ANALYSIS AND MACHINE INTELLIGENCE}%
{Shell \MakeLowercase{\textit{et al.}}: Bare Demo of IEEEtran.cls for Computer Society Journals}
%

\newcommand{\overbar}[1]{\mkern 1.5mu\overline{\mkern-1.5mu#1\mkern-1.5mu}\mkern 1.5mu}
\newcommand{\norm}[1]{\left\lVert#1\right\rVert}



\IEEEtitleabstractindextext{%
\begin{abstract}
    Knowledge graph data are prevalent in real-world applications, and knowledge graph neural networks (KGNNs) are essential techniques for knowledge graph representation learning. Although KGNN effectively models the structural information from knowledge graphs, these frameworks amplify the underlying data bias that leads to discrimination towards certain groups or individuals in resulting applications. Additionally, as existing debiasing approaches mainly focus on the entity-wise bias, eliminating the multi-hop relational bias that pervasively exists in knowledge graphs remains an open question. However, it is very challenging to eliminate relational bias due to the sparsity of the paths that generate the bias and the non-linear proximity structure of knowledge graphs. To tackle the challenges, we propose Fair-KGNN, a KGNN framework that simultaneously alleviates multi-hop bias and preserves the proximity information of entity-to-relation in knowledge graphs. The proposed framework is generalizable to mitigate the relational bias for all types of KGNN. We develop two instances of Fair-KGNN incorporating with two state-of-the-art KGNN models, RGCN and CompGCN, to mitigate gender-occupation and nationality-salary bias. The experiments carried out on three benchmark knowledge graph datasets demonstrate that the Fair-KGNN can effectively mitigate unfair situations during representation learning while preserving the predictive performance of KGNN models.
\end{abstract}

\begin{IEEEkeywords}
    Fairness, Relational Bias, Knowledge Graph Debiasing.
\end{IEEEkeywords}}

\maketitle

\IEEEdisplaynontitleabstractindextext

%
\IEEEpeerreviewmaketitle

\begin{figure}[t!]
\begin{center}
    \includegraphics[width=0.4\textwidth]{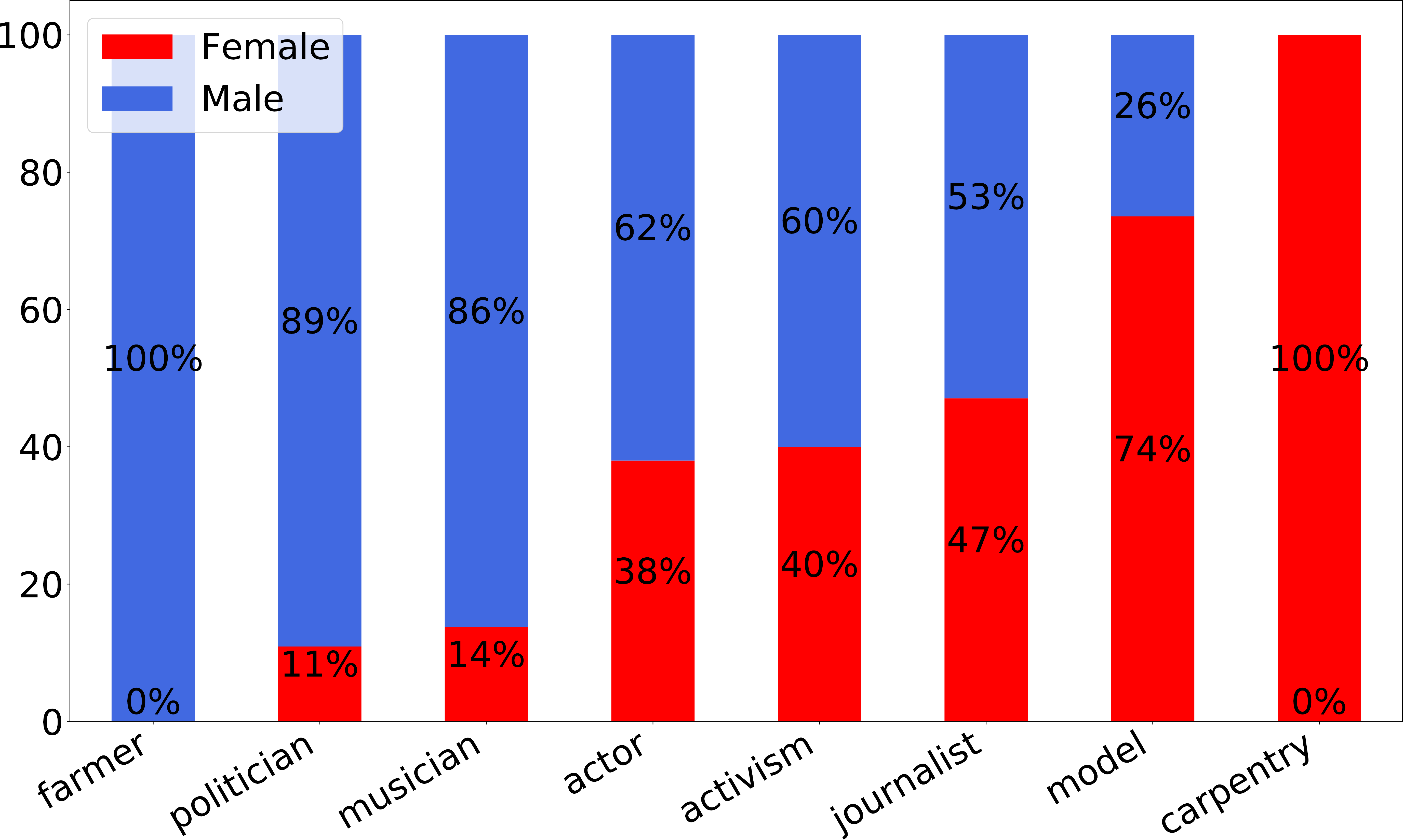}
    \caption{A high-order relational bias example: gender-to-occupation distribution of FreeBase15k-237 dataset.}
\label{fig:histfair}
\end{center}
\end{figure}

\IEEEraisesectionheading{\section{Introduction}\label{sec:introduction}}
\IEEEPARstart{K}{nowledge} graphs are widely adopted to model human knowledge data for many real-world applications, such as recommender systems~\cite{kgrec1,kgrec2}, search engine~\cite{kgss1}, and question-answering systems~\cite{kgqa1}.
Recently, social scientists have warned that discriminative decisions originating from knowledge data bias are negatively impacting our society in both the individual~\cite{cb1,cb2} and the universal~\cite{sp1,sp2} aspects. The data bias may lead to discriminative decisions~\cite{ciampaglia2015computational} and further aggravate human behaviors that lead to more bias in the future data collection process. Data bias exists in knowledge graphs due to the imbalanced data collection~\cite{radstok2021knowledge}. For example, a triplet \texttt{Indian}-\texttt{has\_religion}-\texttt{Hinduism} is collected excessively in the FreeBase15k-237 dataset~\cite{fb15k} (FB15k-237), leading to entity bias in determining whether \texttt{Indian} are \texttt{Hinduism} or not. Moreover, Figure~\ref{fig:histfair} summarizes the high-order relation between genders and occupations in the FB15k-237. There exists an obvious gender-occupation relational bias beyond entity-wise relations.

Existing knowledge graph modeling methods~\cite{transE,transR,complex,chao2020pairre,rotate} preserve the connectivity patterns underlying proximity relation information of individual entities. These models naturally inherit or even amplify the entity bias in the knowledge graph data, leading to discriminatory prediction in downstream applications~\cite{FairGraphMining,agarwal2021towards,gerritse2020bias,fabris2022algorithmic,ramnath2020seeing}. 
To tackle the problem, recent studies alleviate data bias by leveraging adversarial learning~\cite{fairgnn,fairkgeadv1,compfairgraph,nifty} and embedding fine-tuning~\cite{fairkgefine1,fairkgefine2} to debias knowledge graph representations.
However, existing approaches only consider entity-wise bias in knowledge graphs, where the implicit fairness issues in higher-order relations are ignored. 
For example, entity bias, such as people-gender and people-occupation bias in Figure~\ref{fig:histfair}, can be mitigated by existing methods, while multi-hop relational bias, such as occupation-people-gender bias, still pervasively exist in the learned representations. Thus, it is critical to mitigate the multi-hop relational bias in knowledge graphs. 

\begin{figure}[t]
\begin{center}
    \includegraphics[width=0.4\textwidth]{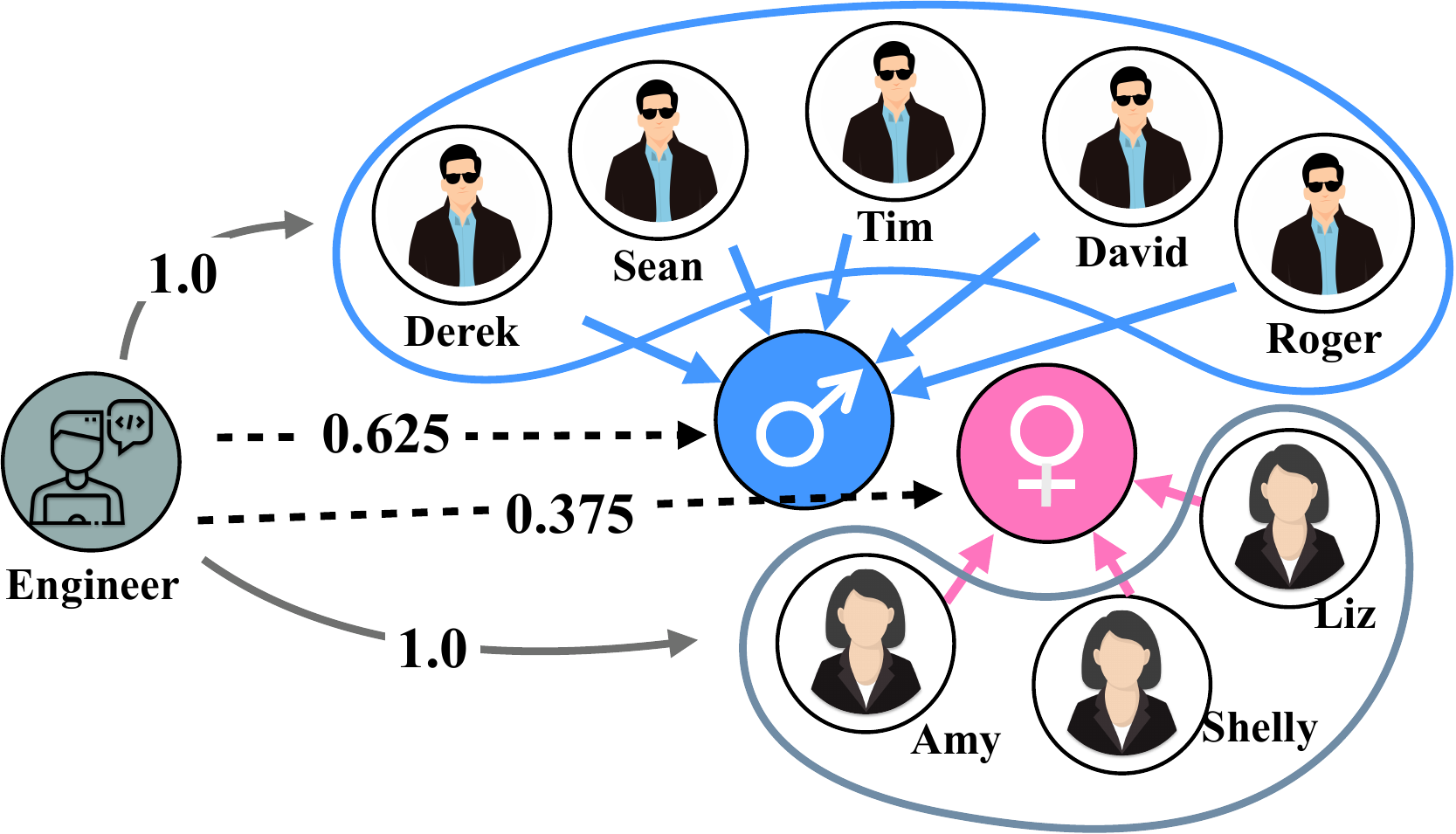}
    \caption{Example of multi-hop relational bias on modeling a knowledge graph. The dot lines denote two-hop gender bias and colored solid lines are the actual relations (e.g., has\_gender and has\_occupation). Due to the imbalanced amount of male and female engineers, the probability of \texttt{Engineer} to \texttt{Male} is 0.625, which is higher than 0.375 of \texttt{Engineer} to \texttt{Female}. The different probabilities of "Engineer" toward male and female cause the multi-hop relational bias.}
    \vspace{-0.1cm}
\label{fig:fairdiag}
\end{center}
\end{figure}

These unfairness issues occur especially in knowledge graph neural networks (KGNN), a crucial technique for knowledge graph representation learning. 
Existing KGNN methods~\cite{rgcn,rgat,compgnn} consider multi-hop information when performing information aggregation through each layer, which may ignore multi-hop relational bias during training. Although KGNN accurately extracts the underlying patterns among entities and relations, data bias is often amplified and leads to unfairness for downstream applications. KGNN thereby suffers from relational bias when considering multi-hop information.
Figure~\ref{fig:fairdiag} illustrates an example of multi-hop relational bias that occurs in KGNN. This type of bias is amplified when modeling the imbalanced distribution of two-hop connections between gender and occupation. For instance, the connectivity probability for "\texttt{Engineer}" to "\texttt{Male}" is higher than "\texttt{Engineer}" to "\texttt{Female}" due to the imbalanced number of people related to the two-gender entities. However, existing entity-wise debiasing methods cannot alleviate the multi-hop relational bias.
For example, to mitigate occupation-gender bias from the multi-hop \texttt{Occupation-People-Gender} path, entity-wise debiasing methods eliminate the bias between "\texttt{People}" and "\texttt{Gender}" by making "\texttt{Gender}" indistinguishable for "\texttt{People}." Although this may implicitly impact the "\texttt{Occupation}" in latent space, the relational bias between "\texttt{Occupation}" and "\texttt{Gender}" remains. The underlying reason is that the unbiased "\texttt{People}" cannot impact the "\texttt{Gender}" due to their indirect relation.

To mitigate the multi-hop relational bias, we aim to develop a fairness knowledge graph modeling framework to alleviate the relational bias and simultaneously preserve the task performance. It is nontrivial to achieve our goal due to the following challenges.
First, relational bias originates from a minority of all multi-hop paths, and it is very challenging to eliminate such minor information. For example, modeling relationships between people naturally lead to numerous multi-hop paths generated from various attributes. However, to eliminate gender-occupation bias, only biased paths that start with gender and end with occupation should be considered for mitigation. 
Second, the elimination of relational bias through existing entity-wise approaches~\cite{fairgnn,compfairgraph,debiaskge,fairkgefine1} may compromise the proximity modeling on bias alleviation. Existing approaches mitigate bias by adopting loss functions such as mean square error in fairness measurement~\cite{bechavod2017penalizing}. Specifically, the proximity structure of a knowledge graph is typically non-linear~\cite{balazevic2019multi}, adopting mean square error arbitrarily assumes the linear relationship between entities. Eliminating bias based on such assumptions will not only hinder relational bias mitigation, but also obfuscate the latent feature space and lead to sub-optimal performances. 

To address the aforementioned challenges, we propose Fair-KGNN, a fairness-oriented framework to mitigate relational bias while maintaining task performance. Specifically, we develop a differentiable bias mitigating framework that can be generally adapted to any existing KGNN method. By adopting the differentiable term, Fair-KGNN can alleviate the multi-hop relational bias while exploiting the multi-hop information simultaneously during the optimization stage. We formally define the unfairness problem with the corresponding two metrics for evaluation. The evaluation of our framework on two state-of-the-art KGNN models addresses two common biases: gender-occupation and nationality-salary bias. The experimental results demonstrate that Fair-KGNN can successfully mitigate the multi-hop data bias on three benchmark datasets while maintaining the performance of downstream tasks. Our contributions can be concluded as follows:

\begin{itemize}[leftmargin=*]
	 \item We detect multi-hop relational bias on knowledge graphs and propose a general framework on KGNN to simultaneously mitigate multi-hop relational bias and preserve high task performance.
	 \item We formally define two fairness metrics to illustrate and evaluate the unfair situations in knowledge graphs representation learning.
	 \item We implement our framework incorporating with two state-of-the-art KGNN models and demonstrate the superiority in fairness improvement and the maintenance of task performance.
\end{itemize} 

\section{Preliminaries} 

In this section, we introduce the KGNNs and the common strategies of fairness preservation in the graphs. Then, we define the multi-hop relational bias problem in knowledge graphs, accompanying our goal of alleviating bias during the knowledge graph representation learning.

\subsection{Knowledge Graph Neural Networks}
Knowledge graph neural networks (KGNN)~\cite{rgnn1,rgnn2,rgnn3,rgnn4,rgnn5,rgcn,rgat,compgnn} aggregate neighborhood information from individual entities by considering relational attributes between each entity. 
Advancements in KGNN mainly focus on developing novel aggregation functions to effectively embed proximity information and multi-hop path information into entity representations.
The relational graph convolutional network (RGCN)~\cite{rgcn} proposed an aggregation function to exploit different relations. Let $r \in \mathcal{R}$ be the relation set, and $\mathcal{N}_{k}(v)$ be the $k$-hop neighborhood of entity $v$, RGCN conducts the relation-wise weighted aggregation of the neighborhood by calculating the following function:
\begin{equation}
    h_{v}^{k} = \sigma \Big( 
    \sum_{r}\sum_{u \in \mathcal{N}_{k}(v)} \frac{1}{|\mathcal{N}_{k}(v)|} 
    \mathcal{W}^{k-1}_{r} h_{u}^{k-1} + \mathcal{W}^{k-1}_{v} h_{v}^{k-1}
    \Big),
\label{eq:rgcn}
\end{equation}
where $\mathcal{W}^{k-1}_{r}$ is a trainable weight matrix for its corresponding relation $r$ in the neighborhood $\mathcal{N}_{k}(v)$, $\mathcal{W}^{k-1}_{v}$ is a trainable parameter for the entity $v$, and $h_{v}^{k}$, $h_{u}^{k-1}$ and $h_{v}^{k-1}$ are latent feature vectors.

Following the success of RGCN, CompGCN~\cite{compgnn} develops a flexible composition operator to jointly aggregate entities with relations in modeling knowledge graphs. CompGCN utilizes the composition operators $\Psi$ to exploit the entity and relation information by updating the following function:
\begin{equation}
    h_{v}^{k} = \sigma \Big( 
    \sum_{(u,r) \in \mathcal{N}_{k}(v)} 
    \mathcal{W}^{k-1}_{\lambda(r)} ~\Psi(h_{u}^{k-1},h_{r}^{k-1})
    \Big),
\label{eq:compgcn}
\end{equation}
where $\mathcal{W}^{k-1}_{\lambda(r)}$ is a trainable weight matrix considering different edge directions (i.e., in, out, and self-loop) for its relation $r$, and $h_{u}^{k-1}$ and $h_{r}^{k-1}$ are the latent feature vectors of the given entity $u$ and relation $r$. The selection of composition operators $\Psi$ formally depends on the different scoring functions adopted.

The level-by-level aggregation mechanism considers only the one-hop neighborhood information, which is therefore difficult to identify multi-hop relational bias. In our work, we consider multi-hop relations between non-sensitive and sensitive entities to mitigate social inclinations, which are the intended behaviors of non-sensitive entities toward certain sensitive entities. We alleviate the information from relation bias while preserve the proximity structure for retaining the model performance. Our framework can be adopted to any existing KGNN method.

\subsection{Preserving Fairness in Graph}

Data bias is widely exhibited in graph structural data and leads to biased prediction results. To alleviate the bias, two approaches have been extensively studied to eliminate the entity-wise bias of representation for the downstream applications: the adversarial training~\cite{fairgnn,debiaskge,adver2,adver3,nifty,compfairgraph,fairkgeadv1}, and the embedding fine-tuning~\cite{fairkgefine1,fairkgefine2}. 

Adversarial training degrades the effects of sensitive entities on the learned feature representations by simultaneously training an accurate predictor against a sensitive entity discriminator. Existing solutions~\cite{compfairgraph,fairkgeadv1} filter the sensitive features embedded in graph representations with an adversarial discriminator. FairGNN~\cite{fairgnn} proposes an ensemble of adversarial debiasing discriminator with the graph convolutional network to alleviate the severe sensitive attributes' bias issues in graph embedding training. The framework adopts adversarial training by minimizing: 

\begin{equation}
    \overbrace{f_C(G, S_a, h_{v})}^{\text{Classifier}} - \overbrace{f_A(S_a, h_{v})}^{\text{Discriminator}},
\label{eq:fairgnn}
\end{equation}

where the classifier $f_C$ aims to learn the entity representations $h_{v}$ for discriminator $f_A$, such that the sensitive attributes of entity $v \in G$ are indistinguishable by $f_A$. 

Embedding fine-tuning post-processes the distribution of the learned representations with heuristic scoring functions to ensure the fairness of entity embeddings.
Existing approach~\cite{fairkgefine1} measures the social bias of gender-related attributes on the learned knowledge graph embeddings (KGE) by designing a scoring function. It computes the difference between two transformation-based distancing scores for people to a sensitive male attribute and a sensitive female attribute. Then, it adjusts the biased distribution of the learned embeddings according to the designed scoring function. Another work~\cite{fairkgefine2} mitigates gender bias in learned KGEs with a debiasing projecting operator to project biased representations in the gender subspace. The projection minimizes the distance between the biased representations and a gender vector to eliminate gender bias in the learned representations.

The two common approaches conduct debiasing on the resulting latent features right before performing the downstream task. However, high-order relational bias remains and leads to unfair consequences. Our framework conducts debiasing by simultaneously considering multi-hop information and sensitive entities during the information aggregation to mitigate both entity and relational bias.

\subsection{Problem Definition}
\sloppy We define the multi-hop fairness problem in knowledge graphs. 
Let $G = ( \mathcal{V}, \mathcal{R})$ be knowledge graphs, where $\mathcal{V} = \left\{ v_1, \dots , v_{|\mathcal{V}|} \right\}$ is the set of entities and $\mathcal{R} = \left\{ r_1, \dots , r_{|\mathcal{R}|} \right\}$ is the set of relations. The sensitive entity set $\mathcal{S}$ collects a sensitive attribute with $q \in \mathbb{N}^{+}$ varieties as its elements and it can be defined as $\mathcal{S} = \left\{ s_1, \dots , s_{q} |~ \forall s_i \in \mathcal{V} \right\}$. For instance, $\mathcal{S}$ can be a set of nationality attributes with Korean, American, and German. The non-sensitive entity set $\mathcal{C}$ collects one kind of non-sensitive attribute with $p \in \mathbb{N}^+$ varieties, which can be defined as $\mathcal{C} = \left\{ c_1, \dots , c_{p} |~ \forall c_j \in \mathcal{V} \setminus\mathcal{S} \right\}$. 
By observing Figure~\ref{fig:fairdiag}, relational bias occurs when a sensitive entity is reachable by a non-sensitive entity, where the connecting paths between the two entities eventually lead to multi-hop relational bias in knowledge graphs. Formally, a biased multi-hop path must include both $(s_i, r_{s})$ and $(r_{c},c_j)$, where $s_i \in \mathcal{S}$, $c_j \in \mathcal{C}$ and $r_{c}, r_{s} \in \mathcal{R}$ denote the relations connecting to the sensitive entity $s_i$ and the non-sensitive entity $c_j$, respectively. Based on the information aggregation of the KGNN models, the biased multi-hop connecting paths $(s_i, r_{s}, \cdots, r_{c}, c_j)$ are exploited hop by hop and therefore lead to the bias of the models in the learned representations. 

Based on the definitions and the intuitions above, we formally define the multi-hop relational bias mitigation problem as follows. Let $\mathcal{P} \equiv (s_i, r_{s}, \cdots, r_{c}, c_{j})$ be a biased multi-hop path in $G$, our goal is to debias the $\mathcal{P}$ by balancing the distribution in the embedding space with a mapping function $f(\mathcal{P}) \rightarrow \mathbb{R}^d$ such that the differences of the correlations between the sensitive entity $s_{i}$ in $\mathcal{P}$ and each $c_{j}$ in $\mathcal{P}$ are minimized while the $r_{s}, r_{c} \in \mathcal{P}$ is preserved. 

\section{Methodology}

\begin{figure*}[!ht]
\begin{center}
    \includegraphics[width=0.9\textwidth]{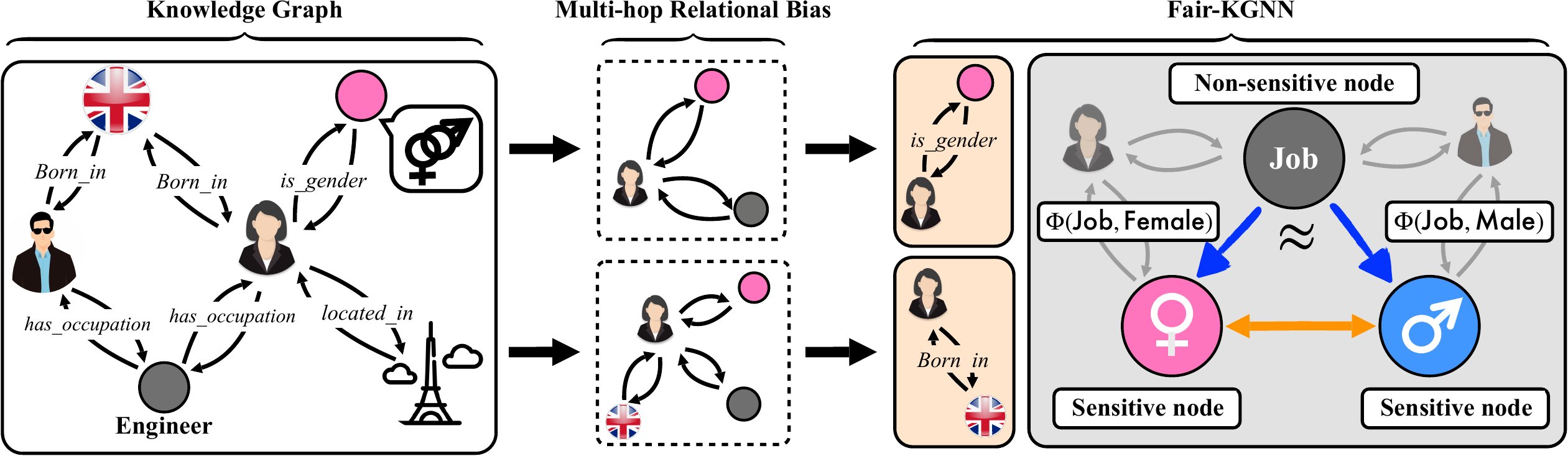}
    \caption{Overview of proposed Fair-KGNN framework. The Fair-KGNN framework contains flexible proximity modeling components (gray part) and two individual fairness components (color maverick part): (1) Sensitive Bias Mitigation (SBM), which is the working flow of the blue arrows, and (2) Entity Relation Preservation (ERP), which represents the orange arrow working flow. SBM tries to mitigate the relational bias (e.g., gender bias) from non-sensitive entity (e.g., job) with measurable function $\Phi(\cdot)$, while ERP ensures every two sensitive attributes are not similar after the representation adjustment from SBM. Afterward, our Fair-KGNN framework are updates based on both fairness components and proximity modeling components.} 
\label{fig:overview}
\end{center}
\end{figure*}

\subsection{Proposed Mitigation Framework}

Our Fair-KGNN framework aims to address relational bias in multi-hop paths while preserving the proximity structure of knowledge graphs into feature vectors. To achieve the goal, we develop two components: SBM and ERP. Figure~\ref{fig:overview} illustrates the framework with two components: SBM (blue arrow) and ERP (orange arrow). 

\noindent\textbf{Sensitive Bias Mitigation (SBM).~} Based on the composition of biased multi-hop paths $\mathcal{P}$, we design Sensitive Bias Mitigation to adjust the relationship between the biased interactions in $\mathcal{P}$ to learn fair representation. Specifically, we aim to equilibrate the inclination of the given non-sensitive entities to sensitive attributes. For example, given a set of multi-hop paths $\mathcal{P}$=\{(\texttt{Engineer}, \textit{has\_occupation}, \texttt{Jack}, \textit{is\_gender}, \texttt{Male}), (\texttt{Nurse}, \textit{has\_occupation}, \texttt{Rose}, \textit{is\_gender} \texttt{Female})\}, we calibrate the non-sensitive entity \texttt{Engineer} to have no social inclination to the two-hop sensitive attributes \texttt{Male} and \texttt{Female} during information aggregation.

Given the sensitive entity set $\mathcal{S}$ in a multi-hop path $\mathcal{P}$ and the non-sensitive entity set $\mathcal{C}$ in $\mathcal{P}$. The objective function of SBM is to mitigate the multi-hop relational bias caused by $\mathcal{P}$, which can achieve by minimizing the inclination between $\mathcal{S}$ and $\mathcal{C}$ with the following loss function:
\begin{equation}
    \arg \min_{\Theta} \sum_{\mathcal{S}} \ln p(\approx_{\mathcal{S}}|\Theta, \mathcal{C}, \mathcal{R}),
\label{eq:fairadj}
\end{equation}
where $\Theta$ is a trainable parameter and $\approx_{\mathcal{S}}$ denotes that every element in the sensitive entity set $\mathcal{S}$ has no inclination toward the non-sensitive entities $c_j \in \mathcal{C}$. 
To compute the inclinations in $\mathcal{P}$, we define the measurable function $\Phi_i(\cdot)$ in $\mathcal{L}_{p}$ space to reveal the inclination scores (IS) between the given non-sensitive entities $c_j \in \mathcal{C}$ with the randomly selected sensitive entity pairs $(s_a, s_b) \in \mathcal{S}$. The IS measures the scores of intended behaviors for non-sensitive entities approaching to sensitive entities.
In different knowledge graph representation learning strategies, the measurable function $\Phi_i(\cdot)$ changes corresponding to the scoring function selections of each representation learning strategy. In this work, we adopt DistMult~\cite{dismult} as the scoring function and thus define our measurable function $\Phi_i(\cdot)$ by following the modeling approach from DistMult:
\begin{align}
    \Phi_{s_i}(h_{c_j}, h_{r_{c}}, h_{r_{s}}, h_{s_i})
    = h_{c_j} \odot h_{r_{c}} \odot h_{r_{s}} \odot h_{s_i},
    \label{eq:fairlike1}
\end{align}
where $r_c$ and $r_s$ mean the relations connecting to $c_j$ and $s_i$, respectively, $h_x$ denotes the latent representation of entity $x$, and
$\odot$ denotes the Hadamard product.
According to statistical parity~\cite{sp} and equal opportunity~\cite{eo}, fairness measurements are composed of every two sensitive scores corresponding to two sensitive entities between other non-sensitive entities. Here, we define
the inclination scores for $(s_a, s_b) \in \mathcal{S}$ by calculating the absolute value between the $\norm{ \Phi_{s_a}(\cdot)}_k$ and $\norm{ \Phi_{s_b}(\cdot)}_k$. We calculate the IS between the non-sensitive entity set and the sensitive entity set, which is equivalent for a non-sensitive entity $c_j \in \mathcal{C}$ to have the same inclination scores for every two sensitive attribute entities $(s_a, s_b) \in \mathcal{S}$. The inclination score (IS) can be formulated as:
\begin{align}
    \text{IS} = |\norm{ \Phi_{s_a}(\cdot)}_k - \norm{ \Phi_{s_b}(\cdot)}_k|,
\end{align}
where we denote $p \in \mathbb{N}^{+}$. 

However, the current IS is not differentiable, which prevents direct optimization of the score. Following the properties from Minkowski inequality~\cite{mikeql}, we ensure that the $\mathcal{L}_{p}$ space is complete normed vector space, which allows optimizing the $p$-norm measurements on $\mathcal{L}_{p}$ space with other measurements on normed vector space simultaneously. Due to the advantages, we follow the Minkowski inequality~\cite{mikeql} to find the upper bound for our current IS as follows:
\begin{align}
    \text{IS} 
    \notag &= |\norm{ \Phi_{s_a}(\cdot)}_k - \norm{ \Phi_{s_b}(\cdot)}_k|
    \leq \norm{ \Phi_{s_a}(\cdot) - \Phi_{s_b}(\cdot)}_k.
\end{align}
The multi-hop relational bias from $\mathcal{P}$ can then be eliminated by minimizing the upper bound term of the IS. 
Therefore, the objective function of SBM can be formulated as follows:
\begin{align}
    \notag & \arg \min_{\Theta} \sum_{\mathcal{S}} \ln p(\approx_{\mathcal{S}}|\Theta, \mathcal{C}, \mathcal{R}) \\
    \notag &=
    \arg \min_{\Theta} \ln \sigma \Big[
    \frac{1}{{\| \mathcal{S} \| \choose 2}}
    \sum_{\substack{c_j \in \mathcal{C}, \\(s_a, s_b) \in \mathcal{S}}} 
    |\norm{ \Phi_{s_a}(\cdot)}_k - \norm{\Phi_{s_b}(\cdot)}_k|
    \Big] \\
    & \equiv
    \arg \min_{\Theta} \ln \sigma \Big[
    \frac{1}{{\| \mathcal{S} \| \choose 2}}
    \sum_{\substack{c_j \in \mathcal{C}, \\(s_a, s_b) \in \mathcal{S}}} \norm{
    \Phi_{s_a}(\cdot) - \Phi_{s_b}(\cdot)
    }_k
    \Big],
    \label{eq:fairlike}
\end{align}
where $\sigma(\cdot)$ is a sigmoid function., $p(\cdot | \cdot) = p(\approx_{\mathcal{S}}|\Theta, \mathcal{C}, \mathcal{R})$ is a likelihood function of our proposed Sensitive Bias Mitigation, $\Phi(\cdot): \mathbb{R}^d \times \mathbb{R}^d \times \mathbb{R}^d \times \mathbb{R}^d \rightarrow \mathbb{R}^d$ is the composition operator and $h_i$ is the feature vectors of the RGNNs for the corresponding nodes $i$.
By minimizing Eq.~\ref{eq:fairlike}, we can minimize IS to alleviate biased multi-hop paths $\mathcal{P}$.

\noindent\textbf{Entity Relation Preservation (ERP).~}
Optimizing Eq.~\ref{eq:fairlike} in SBM disturbs the original embedding distribution of both sensitive entities $s_i$ and non-sensitive entities $c_j$ in $\mathcal{P}$, which may lead to sub-optimal task performance since the proximity structure from $\mathcal{P}$ may not be well preserved in the embedding space. This means that we need to employ the additional embedding adjustments to prevent the information lost of embedding after adopting proposed SBM. To address this problem, we formally propose ERP to enlarge the distance of sensitive attribute entities by maximizing:
\begin{equation}
    \arg \max_{\Theta} \sum_{\mathcal{S}} \ln p(\mathcal{S}| \Theta),
\label{eq:attrclar}
\end{equation}
where $\Theta$ is the trainable parameters. Since Eq.~\ref{eq:fairlike} minimizes the inclination scores pair-by-pair in $\mathcal{S}$, the representations of the sensitive attribute entities $s_i$ are vulnerable to be obfuscated. The obfuscated representations may further lead to indistinguishable representations of the non-sensitive entities $c_j$ when aggregating the information of the obfuscated sensitive attributes to the non-sensitive entities. Thus, we aim to maximize the distance of every two sensitive attribute entities to prevent the representations from being obfuscated. We use the Minkowski distance $\mathcal{D}$ to measure the distance between each two sensitive attribute entities with their corresponding representations. 
The likelihood function of ERP in Eq.~\ref{eq:attrclar} is defined as:

\begin{align}
    p(\mathcal{S}| \Theta)
    \notag &= \sigma \Big[
    \frac{1}{{\| \mathcal{S} \| \choose 2}}
    \sum_{(s_a, s_b) \in \mathcal{S}}
    \mathcal{D}(h_{s_a}, h_{s_b})
    \Big] \\
    &=
    \sigma \Big[
    \frac{1}{{\| \mathcal{S} \| \choose 2}}
    \sum_{(s_a, s_b) \in \mathcal{S}}
    \norm{
    h_{s_a} - h_{s_b} }_k
    \Big],
    \label{eq:clarlike}
\end{align}
where $\sigma(\cdot)$ is a sigmoid function and $h_i$ is the feature vectors of the KGNN for the corresponding sensitive attribute entities $(s_a, s_b)$ in $\mathcal{S}$. We intentionally avoid using the intuitive dot product to discriminate between the two sensitive attributes. The reason is that two sensitive attributes are not connected in the knowledge graphs. Thereby, we follow the same metric space as in Eq.~\ref{eq:fairlike}.

\noindent\textbf{Fair Ratio Loss Function.}
In order to jointly optimize our SBM and ERP with downstream task loss, we integrate the aforementioned objectives denoted as Fair Ratio loss $\mathcal{L}_{\text{FR}}$ by combining the loss functions from Eq.~\ref{eq:fairadj} and Eq.~\ref{eq:attrclar} as follows:
\begin{align}
    \notag \mathcal{L}_{\text{FR}} = & \arg \min_{\Theta} \sum_{\mathcal{S}} 
    \Big(
    \ln p(\approx_{\mathcal{S}}|\Theta, \mathcal{C}, \mathcal{R}) -
    \ln p(\mathcal{S}| \Theta)
    \Big) \\
    \notag = & \arg \min_{\Theta} \sum_{\mathcal{S}} 
    \Big(
    \ln
    \frac{p(\approx_{\mathcal{S}}|\Theta, \mathcal{C}, \mathcal{R})}{
    p(\mathcal{S}| \Theta)}
    \Big) \\
     \propto  & \arg \min_{\Theta} \sum_{\mathcal{S}} 
    \frac{p(\approx_{\mathcal{S}}|\Theta, \mathcal{C}, \mathcal{R})}{
    p(\mathcal{S}| \Theta)}.
\label{eq:fairratio}
\end{align}

To summarize, SBM mitigates the relational bias on the sensitive attribute entities toward each non-sensitive entity. ERP targets enlarging the similarity between every two sensitive attributes obfuscated by SBM. Our Fair Ratio Loss can implicitly normalize the sensitive attributes by groups. This may alleviate the over-smoothing issue caused by deeper KGNN~\cite{deepgnn} and therefore relatively maintain performance on downstream tasks. In the experiment conducted, we present our framework on mitigating gender and nationality bias and set a norm as $k=2$ for computational efficiency.


\subsection{Fair Link Prediction}
We discuss how to incorporate the Fair Ratio loss $\mathcal{L}_{\text{FR}}$ with KGNN to preserve the proximity information while debiasing the multi-hop relational bias in the generated feature representations. As the Fair Ratio loss can be a stand-alone optimization objective, it is capable of debiasing multi-hop relational bias for any kind of downstream tasks. As link prediction is one of the most widely adopted applications in knowledge graphs~\cite{ji2021survey}, we evaluate the effectiveness of our framework on link prediction tasks. It may also be applied to classification tasks.

To perform link prediction, we follow the setting of state-of-the-art KGNN~\cite{rgcn,compgnn}. Specifically, we adopt DisMult~\cite{dismult} as the scoring function, which has been empirically shown to be effective on the standard link prediction benchmark. 
Then, we optimize the cross-entropy loss $\mathcal{L}_{0}$ to model the factual triples with the DistMult scoring function. 
Finally, we incorporate $\mathcal{L}_{0}$ with Fair Ratio loss $\mathcal{L}_{\text{FR}}$ as the final loss function to fair link prediction task. 
By doing this, our proposed Fair Ratio loss helps the KGNN models mitigate the unfair effects while modeling the multi-hop information. 
Specifically, we optimize the proposed Fair-KGNN by illustrating the following joint loss:
\begin{align}
    \mathcal{L} = \mathcal{L}_{0} + \mu \mathcal{L}_{\text{FR}} 
\label{eq:flploss}
\end{align}
where $\mathcal{L}$ denotes the loss to Fair-KGNN and $\mu$ is a weighting parameter to balance the mitigating action for the downstream tasks. By optimizing $\mathcal{L}$, we are able to preserve the link prediction performance while mitigating the multi-hop relational bias during the training procedure. 
 
\subsection{Quantitative Fairness Metrics for Knowledge Graph Representations}
Existing quantitative fairness metrics, statistical parity~\cite{sp} and equal opportunity~\cite{eo}, focus on evaluating the fairness levels of the node classification task. However, we cannot directly apply the two fairness metrics in measuring the fairness levels in link prediction since it is designed for binary classification tasks. As link prediction is also an essential task on knowledge graphs, we modify two evaluation metrics for fair link prediction. 

We evaluate the fairness of link prediction by averaging the statistical parity and equal opportunity of each non-sensitive entity to sensitive attributes. Specifically, during the iteration of each non-sensitive entity $c_j \in \mathcal{C}$, we orderly assume the ground-truth answers when $y = c_j$ and other wrong answers as $y = 0$ with the given sensitive attribute entities $s_i \in \mathcal{S}$.
To leverage the likelihoods of predictions among the sensitive attribute set $\mathcal{S}$, we then calculate the absolute difference between the two likelihoods to generate the fairness scores for each $c_j$ towards every paired sensitive attribute entity in $\mathcal{S}$. 
Note that the computing process goes through all pairwise combinations in $\mathcal{S}$ which is $ M = {\| \mathcal{S} \| \choose 2}$ pairs of sensitive attribute entities.
For example, if we encounter "nationality" as the sensitive attributes set with "Asian", "American", and "European", we have $ M = {3 \choose 2} = 3$ with the pairwise combinations of (Asian, American), (American, European), and (Asian, European).
After computing the absolute difference values among all $c_j \in \mathcal{C}$ throughout each pairwise combination in $\mathcal{S}$, we then average all the fairness scores for each $c_j$ and each pairwise combination of sensitive attributes to receive our final fairness scores. The proposed metrics \textbf{average statistical parity} $\overbar{\Delta_{\text{SP}}}$ and \textbf{average equal opportunity} $\overbar{\Delta_{\text{EO}}}$ can be defined as following equations:

\begin{equation}
   \begin{split}
        \overbar{\Delta_{\text{EO}}} = 
        \frac{1}{M \cdot |\mathcal{C}|} \sum_{\mathcal{S}, \mathcal{C}}
        | P&(\hat{y}_{c_j}=c_j | y=c_j,s_i=0) \\
        &  - P(\hat{y}_{c_j}=c_j | y=c_j,s_i=1) | \\
        \overbar{\Delta_{\text{SP}}} = 
        \frac{1}{M \cdot |\mathcal{C}|} \sum_{\mathcal{S}, \mathcal{C}}
         | P & (\hat{y}_{c_j}=c_j | s_i=0) \\
        & - P(\hat{y}_{c_j}=c_j | s_i=1) |
   \end{split}
\label{eq:avglp}
\end{equation}
where both $\overbar{\Delta_{\text{SP}}}$ and $\overbar{\Delta_{\text{EO}}}$ are evaluated on the given test set and $\hat{y}_{c_j}$ denotes the prediction of the Fair-KGNN. Note that the lower values of $\overbar{\Delta_{\text{SP}}}$ and $\overbar{\Delta_{\text{EO}}}$ denote more fair results in the link prediction task. As for the standard of choosing the non-sensitive attributes and sensitive attributes, we explore the existing definitions~\cite{du2020fairness} for the sensitive attributes in our case, such as gender and nationality.

\section{Experiments}
In this section, we conduct experiments to answer the following four research questions:
\begin{enumerate}[label=\textbf{Q\arabic*},leftmargin=*]
	 \item \label{q1} Compared with the existing baseline methods, can our Fair-KGNN in Eq.~\ref{eq:flploss} alleviate the unfair multi-hop relational bias on knowledge graph representation learning?
	 \item \label{q2} Can proposed Fair Ratio loss in Eq.~\ref{eq:fairratio} help mitigating the multi-hop relational bias and preserving the performance when incorporating with different KGNNs?
	 \item \label{q3} How does Fair Ratio loss impact the inclination scores of the non-sensitive entities toward the sensitive entities?
	 \item \label{q4} How does the hyper-parameters influence the fairness performance of proposed Fair-KGNN?
\end{enumerate}

\subsection{Datasets}~\label{sec:dataset}
\noindent
We test our framework on three publicly available datasets, which include FB15K-237~\cite{fb15k}, Pokec-KG~\cite{pokec} and NBA-players\footnote{https://www.kaggle.com/noahgift/social-power-nba}. The statistical summaries are to the reference of dataset.
In the FB15K-237 dataset, we discuss the fairness problem on the gender-occupation bias, in which gender is treated as the sensitive attribute and occupation is viewed as the non-sensitive attribute. For the Pokec-KG dataset, we also focus on solving the gender-to-occupation fairness analytics. Specifically, the Pokec-KG dataset is sampled with five representative attributes from the original Pokec dataset, and then the selected attribute types are managed as the relations to build a knowledge graph. In NBA-players dataset, we consider nationality-salary bias. We transfer the salary attribute into six different salary groups and gain the nationality attribute with "U.S. players" and "non-U.S. players." Then, we choose all attributes and transfer the attributes as relations to create an NBA-players knowledge graph. Except for the FB15k-237 dataset following the original setting, we split our datasets by taking 70\%, 10\%, and 20\% as the training set, validation set, and testing set.

\subsection{Baselines and Settings}
We compare our Fair-KGNN framework to five representation learning baselines and two debiased KGE models (Adversarial training and Embedding fine-tuning). The hyperparameters are all decided after the grid search. An early stopping strategy is adopted to prevent overfitting. The baseline descriptions are as follows:
\begin{itemize}[leftmargin=*]
    \item {\bf TransE~\cite{transE}} learns the knowledge graph representations via exploiting the translation information from entities to relations.
    \item {\bf DistMult~\cite{dismult}} learns the knowledge graph representations from a bilinear objective model where the structural information is modulized by the multiplication from the head entities through the relations to the tail entities.
    \item {\bf PairRE~\cite{chao2020pairre}} proposes to learn the knowledge graphs representations on complex coordinate space by encoding multiple relation patterns and entities information.
    \item {\bf Social-KGE~\cite{fairkgefine1}} is a fine-tuning-based model which adjusts the pre-trained TransE embeddings based on its own proposed measuring bias scoring function to mitigate the unfair situations after the knowledge graph representation learning.
    \item {\bf Debiasing-KGE~\cite{debiaskge}} develop an adversarial learning framework to prevent biased information learned from the designated biased triplets in knowledge graph representation learning.
\end{itemize}
We also include \textbf{RGCN}~\cite{rgcn} and \textbf{CompGCN}~\cite{compgnn} as our baseline methods to evaluate fairness performance, since we incorporate the Fair-KGNN framework with RGCN and CompGCN to verify the robustness of our framework. We adopt mean reciprocal rank (MRR)~\cite{simple} as the evaluation metric, which is a common metric to reveal the performance of link prediction. All Fair-RGNN results are reported under $\mu=0.7$. Adam optimizer with a learning rate from 0.001 to 0.01, and embedding dimension as 64 are adopted in Fair-RGNN and Baselines.

\begin{table*}[tbh!]
\large
\centering
\makebox[1\textwidth][c]{
\resizebox{1\textwidth}{!}{
\begin{tabular}{l || ccccc||ccccc||ccccc}
\toprule
& \multicolumn{5}{c}{FB15k-237}
& \multicolumn{5}{c}{Pokec-KG} 
& \multicolumn{5}{c}{NBA-players} 
\\
\cmidrule(lr){2-6}\cmidrule(lr){7-11}\cmidrule(lr){12-16}
& MRR & $\overbar{\Delta_{SP}}\blacklozenge$ & $\overbar{\Delta_{EO}}\blacklozenge$ & $\overbar{\Delta_{SP}}\Diamond $ & $\overbar{\Delta_{EO}}\Diamond$ & MRR & $\overbar{\Delta_{SP}}\blacklozenge$ & $\overbar{\Delta_{EO}}\blacklozenge$ & $\overbar{\Delta_{SP}}\Diamond$ & $\overbar{\Delta_{EO}}\Diamond$ & MRR & $\overbar{\Delta_{SP}}\blacklozenge$ & $\overbar{\Delta_{EO}}\blacklozenge$ & $\overbar{\Delta_{SP}}\Diamond$ & $\overbar{\Delta_{EO}}\Diamond$ \\
\midrule
Social-KGE~\cite{fairkgefine1} & 0.1592 & $\dagger$0.0053 & $\dagger$0.0111 &
$\dagger$0.0119 & $\dagger$0.0064 & 0.0696 & 0.0017 & $\dagger$0.0003 & $\dagger$0.0003 & $\dagger$0.0002 & 0.2262 & 0.1071 & $\dagger$0.1152 & 0.0428 & 0.2584\\
Debiasing-KGE~\cite{debiaskge} & 0.1352 & 0.0106 & 0.0420 & 0.0233 & 0.0261 & 0.1220 & 0.1120 & 0.0031 & 0.0590 & 0.0067 & 0.2758 & 0.1142 & 0.1139 & 0.0857 & 0.1514 \\
TransE~\cite{transE} & 0.1605 & 0.0209 & 0.0153 & 0.0124 & 0.0181 & 0.0717 & 0.0696 & 0.0011 & 0.0004 & 0.0012 & 0.2082 & 0.2357 & 0.1435 & 0.2571 & 0.2490\\
DisMult~\cite{dismult} & 0.1507 & 0.0277 & 0.0305 & 0.0398 & 0.1961 & 0.0733 & 0.0017 & 0.0012 & 0.0017 & 0.0172 & 0.2231 & $\dagger$0.0428 & 0.4584 & $\dagger$0.0214 & $\dagger$0.0833 \\
PairRE~\cite{chao2020pairre} & 0.1864 & 0.0022 & 0.0226 & 0.0151 & 0.0061 & 0.1501 & $\dagger$0.0009 & 0.0012 & 0.0004 & 0.0013 & 0.2906 & 0.0286 & 0.2307 & 0.0858 & 0.3043\\
RGCN~\cite{rgcn} & 0.1701 & 0.0271 & 0.0531 & 0.1406 & 0.2492 & 0.0635 & 0.0402 & 0.0121 & 0.0059 & 0.0102 & 0.2471 & 0.0785 & 0.2407 & 0.1571 & 0.4043 \\
CompGCN~\cite{compgnn} & $\dagger$0.2221 & 0.0341 & 0.0343 & 0.0513 & 0.0633 & $\dagger$0.2174 & 0.0221 & 0.0024 & 0.0241 & 0.0027 & $\dagger$0.3489 & 0.1357 & 0.2612 & 0.2500 & 0.2583 \\
\midrule\midrule
Fair-KGNN (RGCN) & 0.1711 & 0.0190 & \bf{0.0038} & 0.0497 & 0.1364 & 0.0665 & 0.0053 & 0.0107 & 0.0028 & 0.0014 & 0.2441 & \bf{0.0253} & \bf{0.0128} & \bf{0.0334} & \bf{0.1217} \\
\midrule
\midrule
Fair-KGNN (CompGCN) & \bf{0.2238} & \bf{0.0003} & 0.0083 & \bf{0.0361} & \bf{0.0312} & \bf{0.2112} & \bf{0.0021} & \bf{0.0003} & \bf{0.0023} & \bf{0.0003} & \bf{0.3460} & 0.0707 & 0.1933 & 0.1255 & 0.2291 \\
\bottomrule
\end{tabular}%
}}
\caption{Performance of fair link prediction on the publicly available datasets, FB15k-237, Pokec-KG, and NBA-players. Recall that lower $\overbar{\Delta_{EO}}$ and $\overbar{\Delta_{SP}}$ are better while larger MRR metric is better. The $\blacklozenge$ denotes the fairness evaluation metrics for the first fairness test, and the $\Diamond$ represents the fairness evaluation metrics for the second fairness test. }
\vspace{0.1cm}
\label{tab:lp2}%
\end{table*}%

\begin{table*}[tbh!]
\centering
\makebox[1\textwidth][c]{
\resizebox{1\textwidth}{!}{
\begin{tabular}{l||c||ccccc||ccccc||ccccc}
\toprule
& \multicolumn{1}{c}{}
& \multicolumn{5}{c}{FB15k-237} 
& \multicolumn{5}{c}{Pokec-KG} 
& \multicolumn{5}{c}{NBA-players} 
\\
\cmidrule(lr){2-2}\cmidrule(lr){3-7}\cmidrule(lr){8-12}\cmidrule(lr){13-17}
& \# Layers & MRR & $\overbar{\Delta_{SP}}\blacklozenge$ & $\overbar{\Delta_{EO}}\blacklozenge$ & $\overbar{\Delta_{SP}}\Diamond$ & $\overbar{\Delta_{EO}}\Diamond$ & MRR & $\overbar{\Delta_{SP}}\blacklozenge$ & $\overbar{\Delta_{EO}}\blacklozenge$ & $\overbar{\Delta_{SP}}\Diamond$ & $\overbar{\Delta_{EO}}\Diamond$ & MRR & $\overbar{\Delta_{SP}}\blacklozenge$ & $\overbar{\Delta_{EO}}\blacklozenge$ & $\overbar{\Delta_{SP}}\Diamond$ & $\overbar{\Delta_{EO}}\Diamond$ \\
\midrule
RGCN & 2 & 0.1751 & 0.0744 & 0.0912 & 0.0801 & 0.2911 & 0.0715 & 0.0875 & 0.0034 & 0.0034 & 0.0028 & 0.2182 & 0.1314 & 0.1895 & 0.1025 & 0.1128 \\
w/ SBM (RGCN) & 2 & 0.1681 & 0.0017 & 0.0553 & 0.0385 & 0.0534 & 0.0583 & 0.0681 & 0.0007 & 0.0040 & 0.0027 & 0.2071 & 0.0460 & 0.0320 & 0.0311 & 0.0846 \\
w/ SBM+ERP (RGCN) & 2 & 0.1729 & 0.0113 & 0.0108 & 0.0438 & 0.1960 & 0.0756 & 0.0021 & 0.0004 & 0.0013 & 0.0015 & 0.2126 & 0.0702 & 0.0626 & 0.0898 & 0.0934 \\
\midrule\midrule
RGCN & 3 & 0.1701 & 0.0271 & 0.0531 & 0.1406 & 0.2492 & 0.0635 & 0.0402 & 0.0121 & 0.0059 & 0.0102 & 0.2471 & 0.0785 & 0.2407 & 0.1571 & 0.4043 \\
w/ SBM (RGCN) & 3 & 0.1642 & 0.0011 & 0.0338 & 0.0178 & 0.2711 & 0.0377 & 0.0041 & 0.0028 & 0.0027 & 0.0066 & 0.2092 & 0.0103 & 0.0366 & 0.0334 & 0.1120 \\
w/ SBM+ERP (RGCN) & 3 & 0.1711 & 0.0190 & 0.0038 & 0.0497 & 0.1364 & 0.0665 & 0.0053 & 0.0106 & 0.0028 & 0.0014 & 0.2441 & 0.0253 & 0.0128 & 0.0334 & 0.1217 \\
\midrule\midrule
CompGCN & 2 & 0.2352 & 0.0241 & 0.0376 & 0.0265 & 0.0253 & 0.2083
 & 0.0422 & 0.0402 & 0.0022 & 0.0450 & 0.3547 & 0.0460 & 0.3082 & 0.0857 & 0.1388 \\
w/ SBM (CompGCN) & 2 & 0.2205 & 0.0145 & 0.0237 & 0.0149 & 0.0155 & 0.2101 & 0.0002 & 0.0001 & 0.0006 & 0.0001 & 0.3541 & 0.0056 & 0.0715 & 0.0172 & 0.1153 \\
w/ SBM+ERP (CompGCN) & 2 & 0.2253 & 0.0059 & 0.0065 & 0.0184 & 0.0202 & 0.2057 & 0.0022 & 0.0011 & 0.0022 & 0.0010 & 0.3526 & 0.0334 & 0.1929 & 0.0403 & 0.1574 \\
\midrule\midrule
CompGCN & 3 & 0.2221 & 0.0341 & 0.0343 & 0.0013 & 0.0633 & 0.2174 & 0.0221 & 0.0024 & 0.0241 & 0.0027 & 0.3489 & 0.1357 & 0.2612 & 0.2500 & 0.2583 \\
w/ SBM (CompGCN) & 3 & 0.2210 & 0.0116 & 0.0110 & 0.0143 & 0.0208 & 0.2119 & 0.0004 & 0.0013 & 0.0005 & 0.0015 & 0.3375 & 0.0637 & 0.1903 & 0.1055 & 0.2091 \\
w/ SBM+ERP (CompGCN) & 3 & 0.2238 & 0.0003 & 0.0083 & 0.0361 & 0.0312 & 0.2112 & 0.0021 & 0.0003 & 0.0023 & 0.0003 & 0.3489 & 0.0707 & 0.1933 & 0.1255 & 0.2291 \\
\bottomrule
\end{tabular}%
}}
\vspace{0.1cm}
\caption{Ablation studies of fair link prediction on Fair-KGNN. The "w/SBM" denotes the Fair-KGNN, only adding SBM (SBM). Fair-KGNN is fully equipped with Sensitive Bias Mitigation and Entity Relation Preservation (ERP) under RGCN and CompGCN, denoted as "w/SBM+ERP." Compared to the baselines, Fair-KGNN with SBM ("w/SBM") obtains the best fairness performance among the three datasets, but the task performance drops. After adding ERP with SBM, the Fair-KGNN equipped with SBM and ERP ("w/SBM+ERP") obtains a similar task performance with the baselines and significantly improves the fairness performance with lower $\overbar{\Delta_{SP}}$ and $\overbar{\Delta_{EO}}$.
}
\label{tab:abl}%
\end{table*}%

\subsection{Quantitative Results}
\subsubsection{\bf{(\ref{q1})} Performance of Fair Link Prediction}
\label{prelim_exp}
\begin{figure}[t]
\begin{center}
    \includegraphics[width=0.45\textwidth]{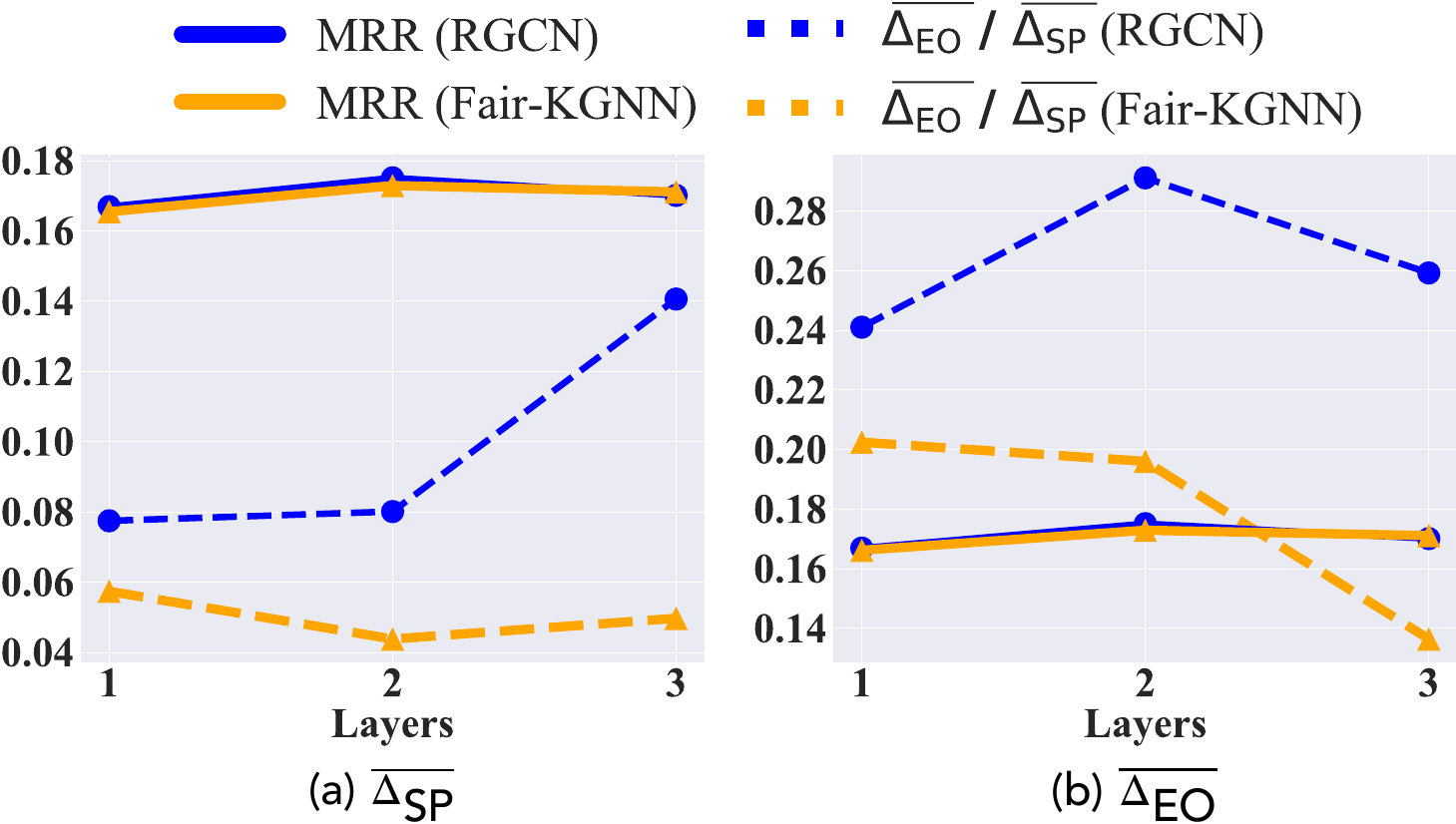}
    \caption{Amplification of the multi-hop relational bias. The multi-hop relational bias amplifies due to the exploitation of multi-hop information. .With the increasing layers in RGCN, we discover that RGCN obtains more multi-hop bias in its learned feature embeddings since RGCN exploits higher hop of paths under larger layers settings.}
\label{fig:prelim}
\end{center}
\vspace{-0.1cm}
\end{figure}

In this section, we provide the results of link prediction using a common evaluation metric: mean reciprocal rank (MRR)~\cite{simple}. Fairness tests are evaluated with two fairness metrics: $\overbar{\Delta_{SP}}$ and $\overbar{\Delta_{EO}}$. We report filtered MRR, which is typically considered to be more reliable~\cite{transE}. 
Before reporting the elimination of multi-hop relational bias, we first need to verify the assumption of whether the data bias increases while the models exploit multi-hop information or not. Figure~\ref{fig:prelim} justify the assumption by illustrating the fairness testing results of the RGCN with different layers. 
We compare RGCN~\cite{rgcn} with 1-layer, 2-layer, and 3-layer. Note that RGCN 1-layer, which does not directly use any multi-hop information, and RGCN 2-layer and RGCN 3-layer, which directly exploit the multi-hop information.
We observe that 2-layer and 3-layer RGCN achieve better performance on the link prediction task than 1-layer RGCN, but the fairness scores of 2-layer and 3-layer RGCN are extremely worse than 1-layer RGCN for two fairness metrics.
The results demonstrate that the cost of performance's improvement with utilizing multi-hop information is embedded a lot of multi-hop relational bias into learned representations.
We also compare RGCN with the proposed Fair-KGNN. Among all the layers settings, Figure~\ref{fig:prelim} shows that Fair-KGNN significantly eliminates the multi-hop relational bias and maintains performance on the link prediction task.

For the fairness test on link prediction, we evaluate our framework on two different tasks of fairness testing. 
The first fairness testing is to evaluate whether the relational bias is eliminated from biased multi-hop paths by Fair-KGNN. Specifically, 
this setting aims to verify that each non-sensitive entity has no bias toward the sensitive attributes. For example, we want to evaluate whether the representations of occupations have gender bias or not. 
The second evaluation of fairness testing performs on the typical one-hop triplets containing non-sensitive entities in knowledge graphs (e.g., people-to-occupation). 
This evaluation aims to test whether the biased information from the sensitive attributes (e.g., people-to-gender) impacts the generating of representations for the non-sensitive entities. The second test verifies the transitivity of multi-hop relational bias during representation learning. For instance, when we sample two connected triplets gender-"Roger"-occupations in KGNN, the gender information embeds into the "Roger's" representation; however, the biased "Roger's" representation leads to the unfair outcomes of predicting the occupations to Roger since it contains biased gender information.
The results of the first and second fairness tests are listed in Table~\ref{tab:lp2}. 

As we can see from Table~\ref{tab:lp2}, Fair-KGNN maintains the performance and significantly mitigates the multi-hop relational bias with the lower $\Delta_{SP}$ and $\Delta_{EO}$ in three publicly available datasets. The best fairness results with the task performance are highlighted in bold. Compared with the debaising baseline methods (Social-KGE and Debiasing-KGE), Fair-KGNN performs better fairness performance with non-degrading task performance. Further key observations are as follows:

First, Fair-KGNN framework is superior to mitigating the multi-hop relational bias. Unlike DisMult, RGCN utilizes deep relational convolution layers to exploit the multi-hop information and embeds a lot of multi-hop relational bias in the learned representations. Moreover, the fairness performance in RGCN, which adopts DisMult as its scoring function, is worse than DisMult. This again reveals the severe situations on amplifying the multi-hop relational bias in KGNN. 
Thus, compared with RGCN and DisMult, our proposed Fair-KGNN framework significantly ameliorates the multi-hop relational bias during the model training. 

Second, Fair-KGNN framework is capable of being incorporated with various KGNN, which utilize the multi-hop information. Specifically, built on RGCN, the Fair-KGNN can relatively maintain the performance of the downstream task and significantly improve the fairness score. We can also observe the same trends for the Fair-KGNN to build upon CompGCN. The results from CompGCN reveal that Fair-KGNN has successfully mitigated the bias of biased data paths by significantly decreasing fairness scores. 


\subsubsection{\bf{(\ref{q2})}~Analytics of Fairness Component Behaviors}
To better understand the effects of the two different components, SBM and ERP, in the proposed Fair-KGNN, we perform ablation studies to analyze and verify their contributions to the Fair-KGNN framework to answer the research question \ref{q2}. The results are listed in Table~\ref{tab:abl}. 
Recall that SBM aims to mitigate the multi-hop relational bias in Fair-KGNN, and ERP is responsible for preserving the downstream task's performance since the adjusted embeddings may impact the performance. 
As shown in Table~\ref{tab:abl}, we observe that SBM works pretty well on mitigating the unfair bias in both RGCN and CompGCN, which makes $\overbar{\Delta_{SP}}$ and $\overbar{\Delta_{EO}}$ lower than that without incorporating SBM. 
We can also discover that the performance of the downstream task drops. SBM ends up with an excellent fairness performance but sacrifices the performance of the downstream task. Thus, ERP is here to deal with the performance dropping issues. As observed in Table~\ref{tab:abl}, the performance of the downstream task reaches back to the same level after adding ERP with SBM. 
This again verifies the necessity of adding two components in Fair-KGNN to ameliorate the unfair multi-hop relational bias and maintain the performance of the downstream task. 

\begin{figure}[t!]
\begin{center}
    \includegraphics[width=0.45\textwidth]{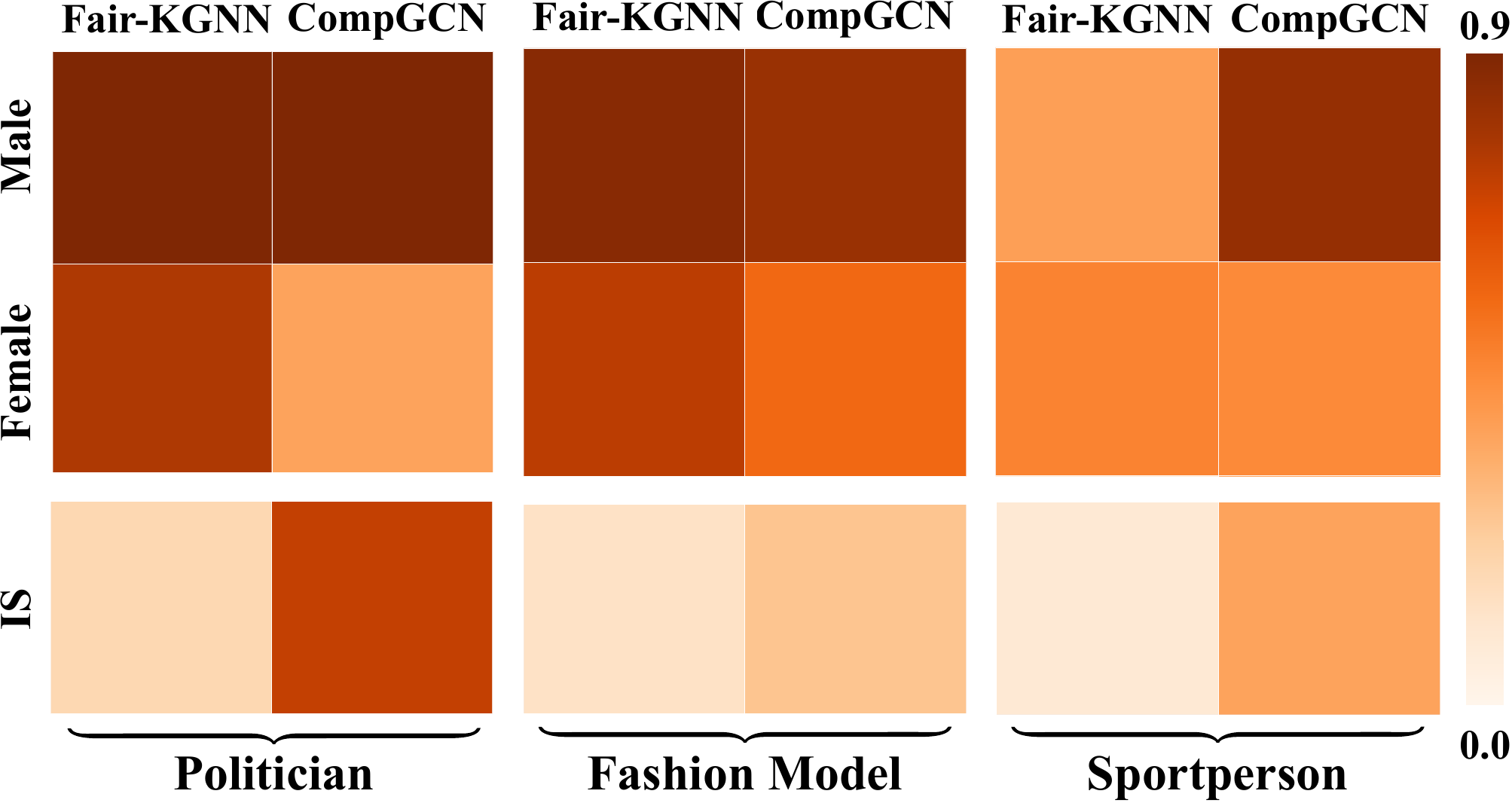}
    \caption{The inclination scores (IS) of the occupations toward gender in the FB15k-237 dataset. The colors represent the individual scores of $\Phi_i(\cdot)$ on three occupations to male and female. The third row represents the IS, where the colors from dark to light represent the severe to slight relational bias situations of the occupations to the gender. We observe that Fair-KGNN achieves better performance on inclination score than CompGCN, which means Fair-KGNN obtains a slighter relational bias than CompGCN.}
\label{fig:dis}
\end{center}
\end{figure}

\subsubsection{\bf{(\ref{q2})}~Analytics of Predictive Performance Behaviors}
In most cases, the task performance in deeper KGNN is lower due to the severe over-smoothing effects. As shown in Table~\ref{tab:abl}, we compare 2-layer RGCN and 3-layer RGCN, and we discover that the task performance for 2-layer RGCN outperforms 3-layer RGCN since 3-layer RGCN may encounter the over-smoothing problem~\cite{deepgnn}. The over-smoothing in RGCN causes its performance to drop from 17.51\% to 17.01\%. Under the impact of the Fair-KGNN framework, some of the task performance increases along with fairness improvement due to the implicit group normalization~\cite{deepgnn} effects provided by ERP. The implicit group normalization eventually alleviates the side effects of over-smoothing.
However, proposed SBM modifies embedding distribution of the biased multi-hop paths in knowledge graphs, which causes slight side effects on task performance. The proposed ERP is here to ensure that the modified attribute entities are not similar and obfuscated, which eliminates the concerns of sub-optimal performance. Thus, the task performance is affected by both SBM and ERP. 

\subsubsection{\bf{(\ref{q3})}~Case Studies of Fairness Pattern}
We analyze the inclination scores of non-sensitive entities toward the sensitive attribute entities to answer the research question \ref{q3}. We choose three occupations, "Fashion Model," "Politician," and "Sportsperson" from the FB15k-237 dataset as the representative for our case studies. "Fashion Model" is a female-biased occupation, while "Politician" and "Sportsperson" are male-biased occupations. 
Figure~\ref{fig:dis} shows the inclination scores of the selected occupations toward two genders from Fair-KGNN and the baseline CompGCN. Each column in the figure reveals the results of computing the inclination scores towards three different occupations. Furthermore, the first row and the second row showcase the output measurements from $\Phi_i(\cdot)$ with respect to male and female, respectively. The last row shows the inclination scores between males and females. For the first two rows, the darker color in the heat map representing the given occupation is more correlated with the given gender based on the output value of $\Phi_i(\cdot)$, while the lighter color reveals the opposite information.

From the last row in Figure~\ref{fig:dis}, we observe that the proposed Fair-KGNN obtains lighter colors than CompGCN among three different occupations. This means that Fair-KGNN acquires lower inclination scores than CompGCN, and Fair-KGNN can effectively eliminate the unfair relational bias during model training. Therefore, Fair-KGNN significantly minimizes the inclination scores for two male-biased occupations and a female-biased occupation. 

\begin{figure}[t]
\begin{center}
    \includegraphics[width=0.45\textwidth]{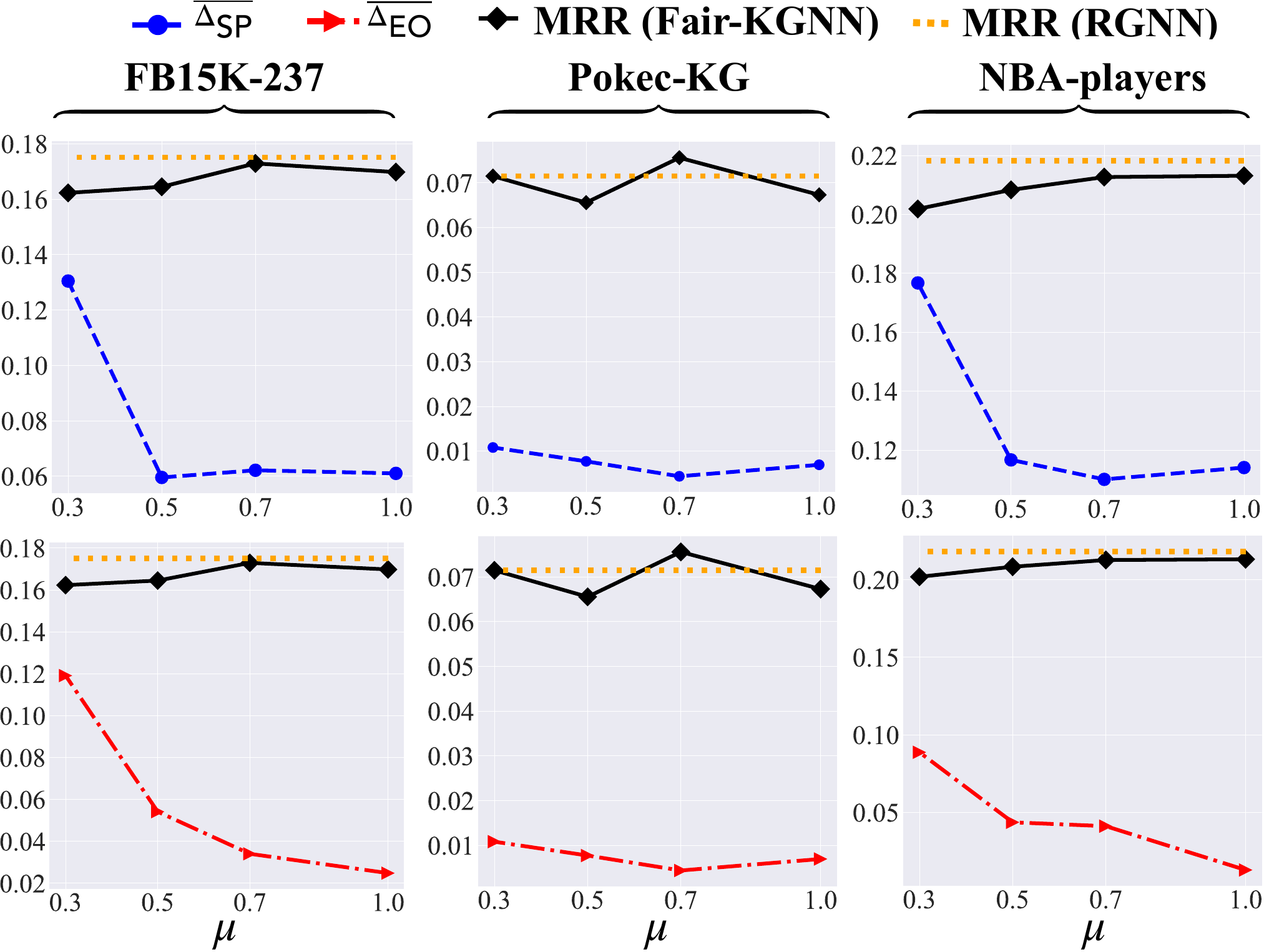}
    \caption{Sensitivity Check on $\mu$ for Fair-KGNN wrt. RGCN. With the similar task performance to the baselines, Fair-KGNN outperforms when the $\mu$ is 0.7 on fairness performance and preserve MRR performance.}
\label{fig:sen}
\end{center}
\end{figure}

\subsubsection{\bf{(\ref{q4})} Analyses of Sensitive Hyperparameter}
In this section, we study the impact of the hyperparameter $\mu$ in Eq.~\ref{eq:flploss} for incorporating Fair-KGNN with 2-layer RGCN to answer the research question \ref{q4}. The hyperparameter $\mu$ controls the balance between the mitigating action and the performance of the downstream tasks. 
Figure~\ref{fig:sen} shows the results of the sensitivity analysis of the hyperparameter $\mu$ with respect to the fairness performance and MRR performance. We evaluate Fair-KGNN by considering $\mu$ from the sets [0.3, 0.5, 0.7, 1.0]. The left column illustrates the results from FB15K-237 dataset, the middle column reveals the results of Pokec-KG dataset, and the right column is for the NBA-players dataset. 

We first observe that the fairness performance is better when the hyperparameter $\mu$ increases; however, the over-adaptation of choosing a larger hyperparameter $\mu$ generally leads to poor fairness performance. As shown in Figure~\ref{fig:sen}, $\mu = 1$ is not always the best choice to mitigate relational bias and preserve MRR performance.
The over adaptation of $\mu$ leads ERP to overly distance each sensitive attribute entity, preventing SBM from minimizing the inclination scores of the non-sensitive entities toward each paired sensitive attribute entity.
Thus, we recommend that $\mu = 0.7$ leads to better fairness results and preserves the task performance better.
Second, in Figure~\ref{fig:sen}, the task performance remains similar to the baselines when Fair-KGNN achieves better fairness scores. The results again demonstrate that Fair-KGNN suffers slightly from the trade-off between fairness performance and task performance. 

\section{Conclusion}
In this paper, we propose Fair-KGNN, a fairness-oriented framework that mitigates the multi-hop relational bias in knowledge graphs while learning the representations. Fair-KGNN is composed of SBM and ERP, eliminating the relational bias from the biased multi-hop paths and preserving the task performance, respectively.
Our work also presents that multi-hop relational bias needs to be considered while alleviating the unfair knowledge graph representations. Experiments on link prediction reveal that Fair-KGNN successfully mitigates unfair relational bias and simultaneously preserves MRR performance. For future directions, based on fair representation learning, we will explore more about fair downstream tasks, such as fair node classification and the recommender systems.

\bibliographystyle{IEEEtranS}
\bibliography{paper}


%

\appendices
\section{Implement Details of Fair-KGNN}\label{apdx:fairr}
To develop Fair-KGNN, we implement the two components of SBM and ERP using Pytorch. In the Fair-KGNN framework, we incorporate Fair-KGNN with two KGNN models, RGCN and CompGCN. As for RGCN implementation, we use one of the most popular open-source packages for graph neural network, PyTorch-geometric~\cite{torchgeo}; for the implementation of CompGCN, we utilize the paper's source codes (\url{https://github.com/malllabiisc/CompGCN}) as the KGNN models to incorporate with our Fair-KGNN framework.

To evaluate the mitigating fairness ability, we test RGCN and CompGCN both on 2-layer settings and 3-layer settings. When training the baseline KGNN model, we utilize the Adam optimizer with a learning rate from 0.001 to 0.01 and a weight decay rate of 0.005 and set the output embedding dimension as 64. We omit the dropout mechanism and ReLU since the RGCN has a strong regularization term to prevent overfitting. 
We find out that the performance is even outstanding if the dropout mechanism and ReLU are omitted. Moreover, according to \cite{nakegnn}, it shows that if the dropout mechanism is omitted, the performance of RGCN is still comparable to the model with the dropout mechanism. However, CompGCN still relies on dropout to prevent overfitting; thus, we keep the dropout mechanism in CompGCN.
Except for the FB15k-237 dataset, we split our dataset by taking 70\%, 10\%, and 20\% as the training set, validation set, and the testing set, respectively. During the data splitting, we also ensure the distribution of the sensitive attributes to be as balanced as possible, which makes our fairness evaluation evaluated objectively and not affected by data distributed bias. The final results are calculated by averaging the results over five repetitions, and all the hyper-parameters in this work are determined under grid search to choose the best performance. Due to limitations of the sensitive attributes on all three datasets, we only consider binary attributes, i.e., female and male as the gender attributes and US players and non-US players as the nationality attributes.

\section{Data Processing Detail}\label{apdx:data}
As described in Section~\ref{sec:dataset}, we choose three publicly available datasets to evaluate Fair-KGNN framework. First of all, in the FB15K-237 dataset, we follow the setting of all the baseline, which has the fixed training set, validation set, and testing set. Second, we get Pokec-KG, a social network dataset from the most popular social network in Slovakia. Since the original data are in the tabular form, we then transfer the tabular data to knowledge graph structural data by choosing the following five attributes: "$\texttt{Age}$," "$\texttt{Gender}$," "$\texttt{Completed\_level\_of\_education}$," "$\texttt{I\_am\_working\_in\_field}$," "$\texttt{Region}$" with respect to "$\texttt{User\_id}$." Last, we select the NBA-players dataset, a statistical record for each NBA player with their performance from 2016-2017. Similar to the preprocessing strategy in the Pokec-KG, we transfer the tabular data into structural graph data and treat the attribute names as the relation in our constructed knowledge graphs. Furthermore, we equally distribute the player's salary into six groups by ensuring each group with nearly equal players.

\begin{table}[h]
\centering
\setlength{\tabcolsep}{5pt}{
\begin{tabular}{l | rrr | rrr}
\toprule
& FB15k-237 & Pokec-KG & NBA-players \\
\midrule 
Entities & 14,541 & 93,245 & 5,138 \\
Relations & 237 & 5 & 99 \\
Train edges & 272,115 & 44,808 & 44,576 \\
Val. edges & 17,535 & 13,104 & 5,024 \\
Test edges & 20,466 & 26,627 & 10,668 \\
Sensitive attribute & Gender & Gender & Nationality \\
\bottomrule
\end{tabular}}
\caption{Dataset statistics for link prediction task.}
\label{tab:datastatic}%
\end{table}%

\section{Proof of Minkowski inequality}~\label{apdx:minkis}
We first reveal the standard version of Minkowski inequality, Let $M$ be the measure space, given $f$ and $g$ be the elements of $\mathcal{L}_p(M)$, where $p \in \mathbb{N}^+$. Due to the closure property of $\mathcal{L}_p(M)$ space, we have $f+g \in \mathcal{L}_p(M)$. Then showing by the triangle inequality, we can have the standard Minkowski inequality to be:
\begin{align}
    \norm{f+g}_p \leq \norm{f}_p + \norm{g}_p,
\end{align}
From the standard Minkowski inequality, we derive that
\begin{align}
    \notag & \norm{f}_p = \norm{(f-g) +g}_p \leq \norm{(f-g)}_p + \norm{g}_p, \\
    \implies & \norm{f}_p - \norm{g}_p \leq \norm{f-g}_p, \label{apdex:eq:1} \\ 
    \implies & \norm{f-g}_p = \norm{g-f}_p \geq \norm{g}_p - \norm{f}_p, \label{apdex:eq:2}
\end{align}
Combining both Eq.~\ref{apdex:eq:1} and Eq.~\ref{apdex:eq:2}, we can have our final inequality:
\begin{align}
    \norm{f-g}_p \geq | \norm{f}_p - \norm{g}_p |_\square.
\end{align}

\section{Baseline Models Details}
\subsection{Implementation Details}
\sloppy For the implementation details of the baseline models, we elaborate the translation-based models and Social-KGE separately.
For translation-based models, we choose to use TransE and Dismult. We select TransE and Dismult as the representative because we want to compare TransE with Social-KGE to see the fairness changes after the fairness adaptation and compare Dismult with RGCN and CompGCN since they select Dismult as the scoring function. For the implementation parts of TransE and Dismult, we use the source code from the RotatE~\cite{rotate} repository, which is a robust implementation toward knowledge graph representation learning research. For the Social-KGE, we implement the codes according to the paper~\cite{fairkgefine1} and adopting by its proposed scoring function to create Social-KGE. For reproducibility, we will share the source code online at an anonymous GitHub (\url{https://anonymous.4open.science/r/Mitigating-Relational-Bias-on-Knowledge-Graphs-B656}) during the reviewing stage.

\subsection{Hyper-parameter Tuning Details}
For the hyper-parameter settings of the baseline models, we select the optimal combination of hyper-parameter by grid search with five repetitions to ensure we get the best and most stable task performance. The hyper-parameters are selected based on the fairness performance, and then we use the same hyper-parameter settings for the fairness test. In the experiments, the regularization term is tested under the set 0.002 to 0.01 with the fixed learning rate 0.0001 and fixed batch size 16. The reported scores are all calculated through the same evaluation process and codes to ensure every metrics score is under control and comparable.



\ifCLASSOPTIONcaptionsoff
  \newpage
\fi

\end{document}